\documentclass[pmlr]{jmlr}

\usepackage{longtable}

\usepackage{booktabs}
\usepackage{cleveref}
\usepackage[load-configurations=version-1]{siunitx} 

\newcommand{\repourl}{\url{https://github.com/MAGNET4Cardiac7T/magnet-preview}}

\theorembodyfont{\upshape}
\theoremheaderfont{\scshape}
\theorempostheader{:}
\theoremsep{\newline}

\jmlrvolume{1}
\jmlryear{2024}
\jmlrworkshop{Machine Learning Meets Differential Equations: From Theory to Applications}

\title[Short Title]{Physical knowledge improves prediction of EM Fields}

\author{
\Name{Andrzej Dulny} \textsuperscript{1} \Email{andrzej.dulny@uni-wuerzburg.de}\\
\Name{Farzad Jabbarigargari} \textsuperscript{2} \Email{Jabbarigar\_F@ukw.de}\\
\Name{Andreas Hotho} \textsuperscript{1} \Email{hotho@informatik.uni-wuerzburg.de}\\
\Name{Laura Maria Schreiber} \textsuperscript{2} \Email{Schreiber\_L@ukw.de}\\
\Name{Maxim Terekhov} \textsuperscript{2} \Email{Terekhov\_M@ukw.de}\\
\Name{Anna Krause} \textsuperscript{1} \Email{anna.krause@informatik.uni-wuerzburg.de}\\
\addr \textsuperscript{1} CAIDAS, University of Würzburg\\
\addr \textsuperscript{2} CHFC, University Hospital Würzburg
}

\editors{Cecília Coelho, Bernd Zimmering, M. Fernanda P. Costa, Luís L. Ferrás, Oliver Niggemann}

\begin{document}

\maketitle

\vspace{-1.2cm}
\section*{Introduction}
\label{sec:intro}
Diagnostic medicine seeks to produce high-quality medical images safely. Ultra-high-field (UHF) MRI improves the signal-to-noise ratio (SNR) over high-field (HF) MRI but introduces safety risks due to increased energy levels in the body, necessitating complex numerical simulations~\citep{caverly_rf_2019}.

Following recent approaches~\citep{gokyar_deep_2023}, we propose using deep learning to replace these simulations by predicting electromagnetic (EM) field distributions with a neural network. Our work enhances 3D models by incorporating physical laws, demonstrating that adding Maxwell's Equations to a 3D U-Net's loss function improves prediction accuracy compared to unconstrained models.
\vspace{-0.2cm}

\section*{Methodology}
\label{sec:methods}


We predict the spatial distribution of EM fields inside the radio-frequency (RF) coil with a subject inside using a 3D U-Net~\citep{cicek_3d_2016}.
We use the phase, amplitude, and positions of RF coils, along with the physical properties: density, permittivity, and conductivity of each point in space, as inputs to our models.
The network is trained using mean squared error to predict the $x, y\text{ and }z$ components of the $E$ and $B$ fields parameterized harmonically as a complex number and separated into the real and imaginary part.
We also train an augmented model, \textit{U-Net Phys}, which incorporates Gauss's law of magnetism ($\nabla\cdot B = 0$) into the loss function using finite differences.

Our data comes from CST Studio Suite simulations of EM fields for an eight-channel dipole array RF coil tuned and matched for 7T MRI (297.2 MHz), with varying subject complexities. We generated 24 simulations, using 19 for training and 5 for testing, each on a $121\times76\times96$ grid.

\begin{table}[ht]
\floatconts
  {tab:results}
  {\caption{Results of our experiments}}
  {\begin{tabular}{lllllllll}
  \toprule
  & \multicolumn{4}{c}{E-Field} & \multicolumn{4}{c}{B-Field}\\
  \cmidrule(r){2-5}\cmidrule(r){6-9}
  & \multicolumn{2}{c}{Subject} & \multicolumn{2}{c}{Total} & \multicolumn{2}{c}{Subject} & \multicolumn{2}{c}{Total}\\
  \cmidrule(r){2-3}\cmidrule(r){4-5}\cmidrule(r){6-7}\cmidrule(r){8-9} 
  \bfseries Model & \bfseries MSE & \bfseries $R^2$ & \bfseries MSE & \bfseries $R^2$ & \bfseries MSE & \bfseries $R^2$ & \bfseries MSE & \bfseries $R^2$ \\
  \midrule
  U-Net & 626.51 & -0.24 & 2341.12 & 0.31 & 0.026 & -1.13 & 0.012 & 0.12 \\
  U-Net Phys & \textbf{276.09} & \textbf{0.46} & \textbf{1623.13} & \textbf{0.52} & \textbf{0.008} & \textbf{0.33} & \textbf{0.007} & \textbf{0.50} \\
  Baseline & 506.71 & 0.00 & 3385.14 & 0.00 & 0.012 & 0.00 & 0.014 & 0.00 \\
  \bottomrule
  \end{tabular}}
\end{table}

\section*{Results and Discussion}
\label{sec:cite}
\Cref{tab:results} presents the results of our preliminary experiments. 
In addition to the plain and physics-augmented U-Net, we include a \textit{mean baseline} model, which always predicts zero for the $E$ and $B$ fields. 
Prediction quality is assessed using mean squared error (MSE) and the coefficient of determination ($R^2$), averaged over the entire domain (Total) and only within the phantom (Subject).

The physics-augmented U-Net (U-Net Phys) significantly outperforms both the plain U-Net and the baseline, particularly in predicting fields inside the subject. 
While the plain U-Net captures the EM field distribution over the whole domain, it struggles inside the phantom, a challenge addressed by the physics-augmented model.

Our findings suggest that incorporating physics-based augmentation into the U-Net architecture leads to more accurate and reliable predictions of the $E$ and $B$ fields. 
In future work we intend to include additional equations, use more simulation data as well as additional neural network architectures to further improve our approach.

We make the code and data used to conduct these experiments publicly available under: \repourl.

\acks{
The project underlying this publication was funded by 
the German Federal Ministry of Education and Research under the grant number
16DKWN0099B (MAGNET4Cardiac7T). 
The responsibility for the content of this publication lies with the authors.
}

\bibliography{bibliography}

\begin{thebibliography}{3}
\providecommand{\natexlab}[1]{#1}
\providecommand{\url}[1]{\texttt{#1}}
\expandafter\ifx\csname urlstyle\endcsname\relax
  \providecommand{\doi}[1]{doi: #1}\else
  \providecommand{\doi}{doi: \begingroup \urlstyle{rm}\Url}\fi

\bibitem[Caverly(2019)]{caverly_rf_2019}
Robert~H. Caverly.
\newblock {RF} {Aspects} of {High}-{Field} {Magnetic} {Resonance} {Imaging} ({HF}-{MRI}): {Recent} {Advances}.
\newblock \emph{IEEE Journal of Electromagnetics, RF and Microwaves in Medicine and Biology}, 3\penalty0 (2):\penalty0 111--119, June 2019.
\newblock ISSN 2469-7257.
\newblock Conference Name: IEEE Journal of Electromagnetics, RF and Microwaves in Medicine and Biology.

\bibitem[Gokyar et~al.(2023)]{gokyar_deep_2023}
Sayim Gokyar et~al.
\newblock Deep learning-based local {SAR} prediction using {B1} maps and structural {MRI} of the head for parallel transmission at 7 {T}.
\newblock \emph{Magnetic Resonance in Medicine}, 90\penalty0 (6):\penalty0 2524--2538, 2023.
\newblock ISSN 1522-2594.

\bibitem[Çiçek et~al.(2016)]{cicek_3d_2016}
Özgün Çiçek et~al.
\newblock {3D} {U}-{Net}: {Learning} {Dense} {Volumetric} {Segmentation} from {Sparse} {Annotation}.
\newblock In Sebastien Ourselin, Leo Joskowicz, Mert~R. Sabuncu, Gozde Unal, and William Wells, editors, \emph{Medical {Image} {Computing} and {Computer}-{Assisted} {Intervention} – {MICCAI} 2016}, pages 424--432, Cham, 2016. Springer International Publishing.

\end{thebibliography}
\end{document}